\documentclass[conference]{IEEEtran} 

\usepackage{preamble}
\usepackage{makecell}
\usepackage{cleveref}
\usepackage[table]{xcolor}

\begin{document}

\title{A Unified XAI-NSAI Analytical Framework for Trustworthy IoT Anomaly Detection}


\author{\IEEEauthorblockN{Anonymous Authors}}

\author{\IEEEauthorblockN{Hugues Leger}
	\IEEEauthorblockA{Department of Computer Science \\
		University of Cincinnati\\
		Cincinnati, OH, USA\\
		legerhs@mail.uc.edu}
\and
\IEEEauthorblockN{Giovani R. Abuaitah}
\IEEEauthorblockA{Department of Computer Science \\
	University of Cincinnati\\
	Cincinnati, OH, USA\\
	abuaitgi@ucmail.uc.edu}
}

\maketitle

\begin{abstract}
	The widespread deployment of Internet of Things (IoT) devices in critical environments makes anomaly detection essential for identifying abnormal behaviors and supporting timely investigation and response. However, accurate detection alone is insufficient. Practical IoT deployments must also provide explanations, incorporate domain knowledge and reasoning, operate under IoT constraints, and support human decision making. Although recently proposed approaches increasingly incorporate Explainable Artificial Intelligence (XAI) or Neuro-Symbolic Artificial Intelligence (NSAI), XAI and NSAI methods are often considered independently. We argue that they should be considered together to better understand their complementary roles in supporting trustworthy IoT anomaly detection. To address this gap, we propose a unified analytical framework organized around five complementary dimensions. The framework provides a structured basis for characterizing existing approaches, identifying capability gaps and design trade-offs, and guiding the design of new anomaly detection systems for different IoT applications. We demonstrate the framework's analytical utility by applying it to three representative studies and illustrate its design utility through its application to a Medical IoT anomaly detection system.
\end{abstract}


\section{Introduction}

Internet of Things (IoT) has become an essential component of modern systems across healthcare, smart infrastructure, industrial automation, transportation, and cyber-physical systems \cite{holzingerCausabilityExplainabilityArtificial2019, abououfExplainableAIEvent2024, nguyenMachineLearningBasedAnomaly2023, mathurosWaXAIExplainableAnomaly2024}. These applications rely on interconnected sensors and devices that continuously collect and exchange data. However, data generated by IoT devices may be affected by sensor faults, device malfunctions, communication errors, environmental disturbances, cyberattacks, or other abnormal conditions, potentially compromising data quality and reliability \cite{abuaitahTaxonomySensorNetwork2015}. Consequently, anomaly detection is essential for identifying abnormal behaviors \cite{abuaitah2015prevalence},  and supporting subsequent investigation and response in IoT environments \cite{ADHIKARI2024100665}.

Recent machine learning (ML) and deep learning (DL) approaches have improved anomaly detection performance in IoT systems \cite{sachdevaAnalysisDeepLearning2022, guptaDeepCNNbasedFramework2023, luRobustAnomalyDetection2024, weiLongshortTermAnomaly2023}. However, many of these models remain difficult to interpret because they operate as black boxes. This lack of transparency can undermine trust in anomaly detection decisions by making it difficult for users to understand why an anomaly was detected and assess whether the decision is justified. This is particularly important in high stakes IoT environments where human decision making remains essential \cite{alaniExplainableEfficientFlowbased2023, gasparExplainableAIIntrusion2024, kalutharageNeurosymbolicLearningXAI2024, fanExplainableAnomalyBasedIntrusion2025}.

Explainable Artificial Intelligence (XAI) addresses this limitation by identifying influential features, temporal patterns, or decision pathways that help users understand why an anomaly was detected \cite{mishraModelExplainabilityInterpretability2022, doEvaluationXAIAlgorithms2024}. However, many XAI methods remain primarily descriptive: they can indicate which features or observations influenced a prediction without establishing whether the detected behavior is consistent with domain knowledge, operational rules, safety constraints, or known attack behavior. Neuro-Symbolic Artificial Intelligence (NSAI) offers a complementary direction by combining data-driven learning with symbolic knowledge and reasoning mechanisms such as rules, logic, ontologies, temporal constraints, and knowledge graphs \cite{yanNeurosymbolicModelsInterpretable2022a, bizzarriNeuroSymbolicArtificialIntelligence2024, kalutharageNeurosymbolicLearningDomain2025, tranNeurosymbolicArtificialIntelligence2025}. By incorporating domain knowledge and explicit reasoning, NSAI can provide stronger semantic grounding and support reasoning beyond feature-level attribution. However, symbolic reasoning alone does not necessarily provide effective human facing explanations of why a data-driven model produced a particular anomaly decision. Thus, combining XAI and NSAI can provide complementary capabilities: XAI helps explain what influenced an anomaly decision, while NSAI helps assess its consistency with domain knowledge and operational constraints. This combination is particularly relevant in IoT environments, where human users must understand, assess, and respond to detected abnormalities.

The presence of both, however, does not by itself establish that they collectively support trustworthy anomaly detection. Their practical value depends on how they are realized within an IoT system, including how and to what extent explanations and reasoning are integrated, how computational constraints are accommodated, and how their outputs support human decision making. For example, a system may provide detailed explanations but lack domain grounded reasoning; incorporate symbolic reasoning but use it only after detection; or tightly connect learning and reasoning while requiring computational, memory, or communication resources that are unsuitable for its deployment environment. Moreover, explanations may have limited practical value if they are not aligned with the role, expertise, or operational needs of the intended user. These examples illustrate that the presence of XAI and NSAI does not adequately characterize how an IoT anomaly detection system supports trustworthiness.

To address this need, we propose an analytical framework for trustworthy IoT anomaly detection systems, derived from a structured analysis of representative research studies. The framework provides a basis for characterizing and comparing existing approaches, identifying missing capabilities and design trade-offs, and enabling researchers to assess the trustworthiness of their own anomaly detection systems across multiple design dimensions.

The main contributions of this paper are as follows: (i) we examine representative IoT anomaly detection research involving XAI, NSAI, and related approaches, and identify a recurring separation of XAI and NSAI methods and/or incomplete consideration of the characteristics relevant to trustworthy anomaly detection. This analysis motivates the need for a multidimensional analytical perspective that examines these characteristics jointly; (ii) we develop an analytical framework for characterizing the trustworthiness of IoT anomaly detection approaches across five design dimensions. The framework provides a structured basis for identifying capability gaps, characterizing design trade-offs, and examining how design considerations are addressed within an anomaly detection system; and (iii) we demonstrate the framework's analytical utility through three representative approaches and its design utility through an illustrative neuro-symbolic healthcare time-series application, showing how the framework can support both comparative analysis and the design of systems for specific applications. Based on these findings, we identify key design considerations and research directions for developing more integrated and practical approaches to trustworthy IoT anomaly detection.

The remainder of this paper is organized as follows.  Section~\ref{sec:foundations} summarizes XAI and NSAI concepts used in the paper. Section~\ref{sec:related_work} reviews related work and positions our paper. Section~\ref{sec:methodology} describes the proposed framework development and application methodology.  Section~\ref{sec:unified_xai_nsai_framework} introduces the proposed framework and outlines its design dimensions.
 Section~\ref{sec:framework_application} details the framework application process and illustrative mapping. Section~\ref{sec:discussion} discusses framework implications and Section~\ref{sec:validity} presents any threats to validity. Section~\ref{sec:conclusion} concludes the paper.

\section{Background}
\label{sec:foundations}

Explainable Artificial Intelligence (XAI) aims to make model decisions understandable by showing which features, signals, time steps, or decision paths influenced a prediction. In IoT anomaly detection, XAI can help users validate alerts, investigate abnormal events, and understand why a system classified a behavior as anomalous. For example, an XAI method may identify an unusual increase in motor vibration and temperature as the primary factors contributing to an anomaly alert, helping an engineer understand which sensor observations influenced the decision. Common explanation forms include local explanations for individual alerts, global explanations of overall model behavior, and concept-level explanations that connect model outputs to domain concepts \cite{mishraModelExplainabilityInterpretability2022, doEvaluationXAIAlgorithms2024}.

Neuro-Symbolic Artificial Intelligence (NSAI) combines data-driven learning with symbolic knowledge and reasoning mechanisms such as rules, logic, ontologies, temporal constraints, and knowledge graphs. In anomaly detection, NSAI can help determine whether an alert violates operational rules, matches known attack behavior, conflicts with physical constraints, or reflects a meaningful system state. For example, symbolic reasoning could combine elevated vibration and temperature with an operational rule indicating a potential bearing fault, thereby interpreting the anomaly in the context of domain knowledge. The degree of symbolic integration varies across systems, ranging from post-detection validation to deeper integration of symbolic knowledge into learning or inference \cite{yanNeurosymbolicModelsInterpretable2022a, bizzarriNeuroSymbolicArtificialIntelligence2024, kalutharageNeurosymbolicLearningXAI2024, kalutharageNeurosymbolicLearningDomain2025}.

 \section{Related Work and Positioning}
\label{sec:related_work}

Existing research relevant to this work spans five areas: trustworthy anomaly detection, explainable IoT anomaly detection, deployment-aware XAI for IoT, socio-technical trust in IoT systems, and neuro-symbolic anomaly detection.

The first area broadly studies trustworthy anomaly detection. Yuan and Wu \cite{yuanTrustworthyAnomalyDetection2022} surveyed trustworthy anomaly detection from the perspectives of interpretability, fairness, robustness, and privacy preservation, showing that trustworthiness requires more than detection performance alone. Bougaham and Fr{'e}nay \cite{bougahamTrustworthyAnomalyDetection2025} proposed tapAUC, a method designed for critical applications where asymmetric error costs make false negatives especially harmful. These works show that trustworthy anomaly detection involves multiple technical and operational considerations beyond predictive accuracy.

The second area focuses on explainable anomaly and intrusion detection using XAI methods such as SHapley Additive exPlanations (SHAP) \cite{SHAP}, Local Interpretable Model agnostic Explanations (LIME) ~\cite{LIME}, feature attribution, surrogate models, and intrinsically interpretable classifiers. These methods improve transparency by helping users identify which features, observations, or time steps contributed to an anomaly decision. Recent publications have applied such approaches to IoT anomaly detection to improve the interpretability of otherwise complex machine learning and deep learning models \cite{alaniExplainableEfficientFlowbased2023, gasparExplainableAIIntrusion2024, fanExplainableAnomalyBasedIntrusion2025}. However, XAI methods primarily focus on explaining model behavior and do not necessarily assess anomaly decisions against explicit domain knowledge, rules, or constraints. This limitation motivates the use of symbolic reasoning as a complementary capability.

The third area examines deployment-aware XAI for IoT anomaly detection. IoT environments impose practical constraints related to computational resources, memory, latency, communication overhead, and the placement of detection and explanation components. Rathi and Shandilya \cite{rathiTrustworthyIoTModular2025} proposed a modular XAI-based framework for real-time IoT anomaly detection at the edge, combining lightweight models with SHAP~\cite{SHAP} and LIME~\cite{LIME} explanations and evaluating deployment-oriented metrics such as inference time, explanation latency, memory usage, and CPU load. This work demonstrates the importance of considering explanation and detection costs together with the resource constraints of IoT environments. More broadly, deployment-aware approaches highlight that explainability mechanisms must be considered in relation to where and how anomaly detection systems operate. However, deployment considerations alone do not establish how symbolic reasoning, explanation mechanisms, and human decision support should be integrated within an IoT anomaly detection system.

The fourth area examines trustworthy IoT anomaly detection from socio-technical and human-centered perspectives. Zeng et al. \cite{zengConceptualFrameworkAIDriven2024} proposed a conceptual framework for AI-driven anomaly detection in IoT networks for smart cities, emphasizing technological readiness, human factors, and socio-technical adoption. Their work shows that trustworthy anomaly detection depends not only on detection capability, but also on human interaction, organizational readiness, and deployment acceptance. These perspectives reinforce the importance of considering the intended users and operational context when evaluating AI based anomaly detection systems.

The fifth area examines neuro-symbolic AI for IoT, cybersecurity, and anomaly detection. Prior work has shown that symbolic rules, logic constraints, ontologies, knowledge graphs, and hybrid reasoning mechanisms can improve interpretability, reasoning consistency, and robustness in AI based detection systems \cite{yanNeurosymbolicModelsInterpretable2022a, bizzarriNeuroSymbolicArtificialIntelligence2024, bohneNeuroSymbolicApproachAnomaly2023, kalutharageNeurosymbolicLearningXAI2024, kalutharageNeurosymbolicLearningDomain2025, tranNeurosymbolicArtificialIntelligence2025}. These papers demonstrate the potential of NSAI to move beyond feature-level explanation toward reasoning informed by domain knowledge. For example, Bizzarri et al. \cite{bizzarriNeuroSymbolicArtificialIntelligence2024} investigated the integration of first-order logic constraints into neural intrusion detection, while Kalutharage et al. \cite{kalutharageNeurosymbolicLearningDomain2025} explored the use of cybersecurity knowledge and knowledge-graph reasoning to support anomaly interpretation. Such studies demonstrate how symbolic knowledge can complement data-driven detection and provide stronger semantic grounding.

Prior work demonstrates the importance of explainability, reasoning based on domain knowledge, deployment constraints, and human factors in trustworthy IoT anomaly detection. However, these aspects are often addressed independently, making it difficult to examine their interactions and, in particular, how XAI and NSAI can jointly contri	bute to trustworthy detection. Our work addresses this gap by bringing these perspectives together within a unified analytical framework.
    
\section{Methodology}
\label{sec:methodology}

A structured literature analysis was conducted to inform the development of the proposed framework. The analysis examined recent research on explainable and neuro-symbolic anomaly detection in IoT and related cyber-physical environments to identify recurring design patterns, capabilities, and limitations relevant to trustworthy anomaly detection.

The analysis considered publications from January 2022 to April 2026. Relevant literature was identified through searches of IEEE Xplore, ACM Digital Library, SpringerLink, ScienceDirect, MDPI Sensors, Wiley Online Library, Google Scholar, and arXiv.org. Search queries combined terms related to explainable artificial intelligence, neuro-symbolic artificial intelligence, anomaly detection, intrusion detection, event detection, fault detection, IoT, Industrial IoT, cyber-physical systems, knowledge graphs, ontologies, rules, and logic. Backward and forward reference checking was also used to identify additional relevant work.

Relevant work was considered when it addressed anomaly detection in IoT, Industrial IoT, Medical IoT, sensor networks, industrial control systems, or related cyber-physical environments, and incorporated at least one of the following: an explainability mechanism, a symbolic reasoning component, or a hybrid XAI--NSAI design. Preference was given to studies that provided empirical validation, a clearly described system architecture, or a methodological contribution that could be analyzed in relation to trustworthy IoT anomaly detection. Research works without a clear connection to the target IoT application domains or without an analyzable explanation or reasoning component were excluded from the detailed analysis.

The initial search identified approximately 160 research records. After duplicate removal and preliminary relevance screening, 120 records remained for title and abstract screening. Full-text assessment resulted in a corpus of 54 publications for further analysis. This corpus was examined to identify recurring characteristics and design patterns across explainability, symbolic reasoning, system architecture, deployment, and human-centered support. Rather than treating the frequency of individual techniques as the primary outcome, the analysis focused on how these characteristics were realized and related within existing anomaly detection systems.

\begin{figure*}[!t]
	\centering
	\includegraphics[
	width=\linewidth]{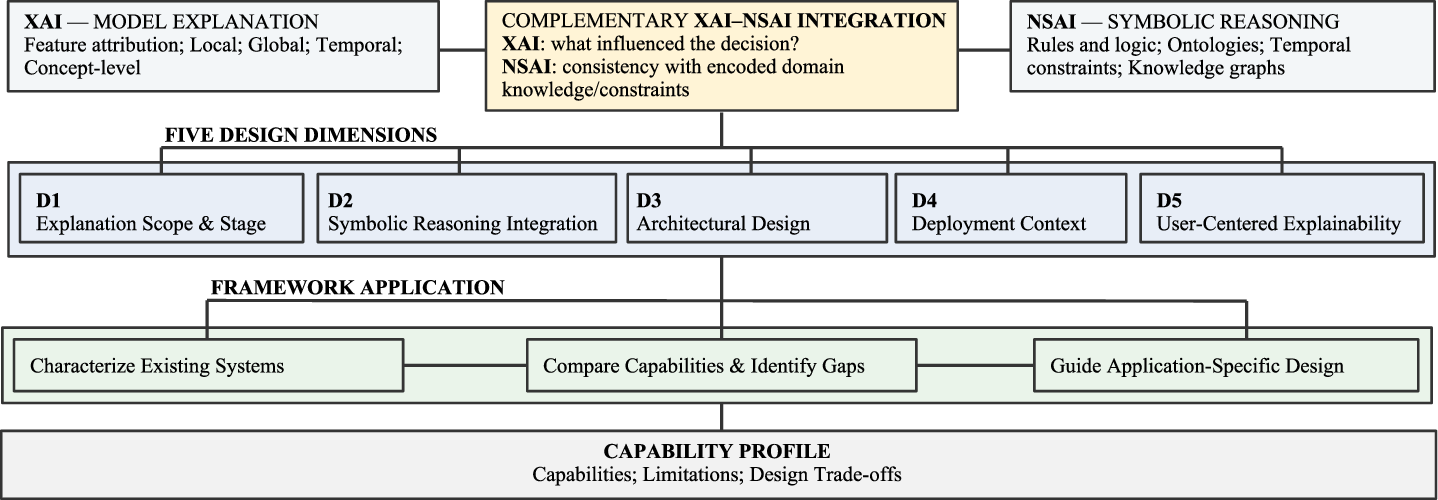}
	\vspace{-5mm}	
	\caption{Proposed unified XAI--NSAI framework for trustworthy IoT anomaly detection.}
	\label{fig:xai_nsai_unified_framework}
	\vspace{-2mm}	
\end{figure*}

For detailed framework analysis, 16 representative models were selected from this corpus. The selection captured variation in how XAI and NSAI capabilities are incorporated, including approaches based on XAI alone, symbolic rules, explicit domain knowledge, and integrated neuro-symbolic methods. It also covered a range of IoT operational settings, including network intrusion detection, Industrial IoT, Medical IoT, industrial control systems, and cyber-physical fault diagnosis.

For each selected work, information was extracted on the application environment, data modality, anomaly detection task, learning architecture, explanation mechanism, symbolic reasoning component, deployment setting, and user-centered explainability features. These characteristics were examined comparatively to identify recurring design considerations, limitations, and trade-offs. The resulting observations informed the formulation of the five analytical dimensions that constitute the framework presented in  Section~\ref{sec:unified_xai_nsai_framework}.

From these 16 models, three representative studies were selected for detailed illustrative mapping because they exemplify distinct design patterns: XAI designed for deployment under IoT constraints, XAI--NSAI approaches that incorporate explicit domain knowledge, and neuro-symbolic approaches that integrate symbolic knowledge directly into learning or inference. This illustrative application demonstrates how the framework can characterize existing approaches, identify capabilities that are addressed only partially or independently, and examine design trade-offs across the proposed dimensions.
   
\section{Proposed Framework}
\label{sec:unified_xai_nsai_framework}

We propose a unified XAI-NSAI analytical framework which provides a structured way to assess how XAI and NSAI jointly support trustworthy IoT anomaly detection. It is intended as a conceptual and analytical framework rather than a deployable system architecture. Its purpose is to compare existing approaches, diagnose missing capabilities, and guide future system design by examining how explanation, reasoning, architecture, deployment, and user needs interact in IoT environments.

\Cref{fig:xai_nsai_unified_framework} presents the conceptual structure of the proposed unified XAI--NSAI framework. Rather than treating XAI and NSAI as separate sets of techniques, the framework organizes their complementary capabilities across five design dimensions: \textit{explanation scope and stage}, \textit{symbolic reasoning integration}, \textit{architectural design}, \textit{deployment context}, and \textit{user-centered explainability}.


The framework yields a multidimensional capability profile that characterizes how an anomaly detection system addresses the five design dimensions and highlights corresponding strengths, limitations, and design gaps.

\subsection{Design Dimensions}
\label{sec:unified_framework_dimensions_design_components}

The five design dimensions shown in \Cref{fig:xai_nsai_unified_framework} establish the analytical structure of the proposed framework and are intended to be considered as complementary design considerations rather than a rigid taxonomy. Table~\ref{tab:unified_framework_dimensions_definition} defines each dimension and its analytical focus.

\begin{table*}[!t]
   	\caption{Design Dimensions of the Proposed Unified XAI--NSAI Framework}
   	\label{tab:unified_framework_dimensions_definition}
  	\centering
\renewcommand{\arraystretch}{1.2}
\begin{tabular}{
		|>{\fontsize{7}{8}\selectfont\RaggedRight\arraybackslash}p{5cm}|
		>{\fontsize{7}{8}\selectfont\RaggedRight\arraybackslash}p{4.4cm}|
		>{\fontsize{7}{8}\selectfont\RaggedRight\arraybackslash}p{8.0cm}|
	}
			
        \hline
        \rowcolor{gray!25}
		\textbf{Dimension} & \textbf{Key Question} & \textbf{Representative Scope}\\
		\hline
		\textbf{D1 \newline Explanation Scope and Stage}
		& What is explained, and when?
		& Intrinsic/ante-hoc; post-hoc; local/global; concept-level; temporal; hybrid\\
		\hline
		\textbf{D2 \newline Symbolic Reasoning Integration} 
		 & How is symbolic knowledge and reasoning incorporated?
		 & None; rule-based; knowledge-based; integration with learning, inference, or decision making\\ 
		 \hline
		\textbf{D3 \newline Architectural Design} 
		& How are detection, explanation, and reasoning organized?
		& Learning-only; pipeline-based; hybrid; tightly integrated\\ 
		\hline
		\textbf{D4 \newline Deployment Context }
		&  Where are detection, explanation, and reasoning executed?
		&  Centralized/cloud; edge; edge-cloud; federated/distributed\\
		\hline
		\textbf{D5 \newline User-Centered Explainability}
		& How do explanations support intended users?
		& Technical; operational; domain-specific; role-, expertise-, task-, or feedback-aware\\
		\hline
	\end{tabular}
	\vspace{-2mm}	
\end{table*}		

The \textbf{explanation scope and stage} dimension (D1) characterizes when explanations are generated and the level at which model behavior is explained. \textit{Intrinsic} or \textit{ante-hoc} explanations incorporate interpretability into the model or its design, whereas \textit{post-hoc} explanations are generated after prediction. \textit{Local} explanations address individual predictions, while \textit{global} explanations characterize broader model behavior. \textit{Concept-level} and \textit{temporal} explanations describe behavior through human understandable concepts or across time, respectively. \textit{Hybrid} approaches combine multiple explanation forms. The \textbf{symbolic reasoning integration} dimension (D2) examines the role and depth of symbolic knowledge and reasoning, ranging from no symbolic component to rule- or knowledge-based reasoning to tighter integration with learning, inference, or decision making. The \textbf{architectural design} dimension (D3) captures how anomaly detection, explanation, and symbolic reasoning components are organized and how they interact within the architecture.\textit{ Learning-only} architectures provide detection without a separate explanation or reasoning component. In a \textit{pipeline-based} architecture, detection, explanation, and reasoning operate as sequential components, whereas in a \textit{hybrid} architecture, learning and symbolic components are combined while retaining distinct functions. In a \textit{tightly integrated} architecture, learning and reasoning interact with the model or inference process. The \textbf{deployment context} dimension (D4) considers where these components are executed. In a \textit{centralized/cloud} deployment, processing occurs at a centralized or cloud resource, while in \textit{edge} deployments, processing occurs on or near a data source. In \textit{edge-cloud} deployments, processing is divided between edge and cloud resources. In \textit{distributed/federated} deployments, processing or learning is coordinated across multiple nodes without relying on a single processing location. Finally, the \textbf{user-centered explainability} dimension (D5) considers how explanations support their intended users. This includes \textit{technical users}, such as developers or technical analysts, who may require detailed information about system behavior; \textit{operational users}, who engage in monitoring or response activities and may require explanations that support timely decision making; and \textit{domain-specific users}, who may require explanations expressed using concepts and terminology meaningful within the application domain. Explanations may also be \textit{adapted} to user role, expertise, task, or feedback. The five dimensions provide a framework-based capability profile for characterizing a system's capabilities, limitations, and design trade-offs. The framework does not prescribe a single preferred architecture or equate any individual design choice with trustworthiness; rather, it supports systematic comparison and helps identify gaps in explanation, symbolic reasoning, deployment, and user-centered support.

\section{Framework Application and Illustrative Mapping}
\label{sec:framework_application}
This section demonstrates how the proposed framework can characterize representative IoT anomaly detection systems, identify capability gaps and design trade-offs, and inform the design of a new anomaly detection system for Medical IoT applications. The mapping is illustrative, not a comprehensive assessment of existing or new approaches.

\subsection{Framework Application Process}

\begin{figure}[!t]
	\centering
	\centerline{\includegraphics[width=\linewidth]{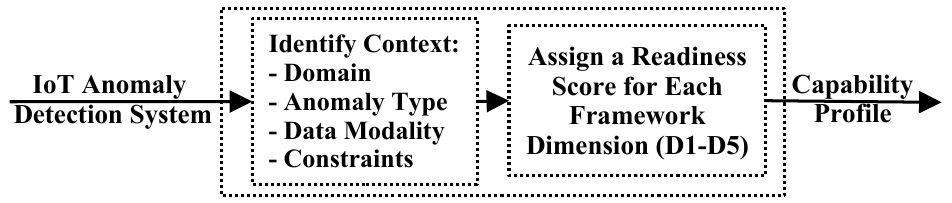}}
	\caption{Application process of the proposed XAI--NSAI framework.}
	\label{fig:framework_application_process}
	\vspace{-2mm}	
\end{figure}

\Cref{fig:framework_application_process} depicts the application process of the proposed framework. To apply the framework, an IoT anomaly detection system is first characterized by its application context, including the target domain (e.g., healthcare, smart homes, industrial automation, transportation, or digital agriculture), anomaly type (e.g., sensor faults, abnormal behavior, intrusions, or system failures), data modality (e.g., time-series sensor data, network traffic, logs, or multimodal data), and deployment constraints (e.g., latency requirements, computational resources, or communication limitations). This contextual characterization establishes the operational setting in which the system is evaluated and provides the basis for interpreting the framework dimensions relative to the intended application environment and deployment requirements. Afterward, each framework dimension (D1--D5) is assessed and an optional qualitative \textit{readiness score} is assigned. The resulting assessment is summarized as an optional framework based capability profile that characterizes the system's strengths, limitations, and design trade-offs across the five dimensions.

\subsection{Framework Readiness Scoring}

To make the framework easier to apply, each dimension (D1--D5)  can be assigned an optional qualitative \textit{readiness score} ranging from 0 to 3. The score is not intended to be a validated quantitative trustworthiness metric; instead, it provides a compact way to summarize the level of support a system provides for each framework dimension. For each dimension, the assigned score reflects the extent to which the system supports the representative scopes listed in Table~\ref{tab:unified_framework_dimensions_definition}. A score of 0 indicates that the corresponding capability is absent or not reported, a score of 1 indicates limited or weak support, a score of 2 indicates moderate support, and a score of 3 indicates strong support.

For a given system, the five dimension scores form an optional qualitative capability profile:
\begin{equation}
	P=(D_1,D_2,D_3,D_4,D_5),
\end{equation}
where $D_1$ through $D_5$ correspond to the five framework dimensions defined in Table~\ref{tab:unified_framework_dimensions_definition}.

\subsection{Framework Application To Representative Studies}
Table~\ref{tab:illustrative_trustworthiness_profile} applies the framework to three  representative studies selected to showcase different levels of XAI--NSAI integration: a post-hoc XAI pipeline, a knowledge grounded XAI--NSAI approach, and an integrated neuro-symbolic model. Rathi and Shandilya~\cite{rathiTrustworthyIoTModular2025} provide a modular edge-oriented XAI pipeline using XGBoost, Isolation Forest, Autoencoders, SHAP, and LIME. The system is strong in deployment-aware explainability and reports latency and resource metrics, but lacks symbolic reasoning, knowledge-based inference, and deeper user-adaptive explanation support. Kalutharage et al.~\cite{kalutharageNeurosymbolicLearningDomain2025} combine autoencoder based detection, SHAP explanations, cybersecurity knowledge graph reasoning, CIA mapping, and MITRE ATT\&CK alignment. The system provides strong domain-grounded interpretation and edge-cloud awareness, but symbolic reasoning is mainly applied after detection. Bizzarri et al.~\cite{bizzarriNeuroSymbolicArtificialIntelligence2024} integrate first-order logic constraints into neural IDS training through a Logic Tensor Network. The system provides strong neural-symbolic integration, but limited user-facing explanation and no demonstrated edge deployment.

Applying our proposed framework to these studies demonstrates its value in revealing complementary strengths and gaps across different design dimensions that may remain obscured when approaches are evaluated primarily by detection performance or individual criteria.

\subsection{Key Patterns Observed Across the Reviewed Studies}

Across the reviewed literature, several recurring patterns were observed. Post-hoc XAI pipelines based on methods such as SHAP, LIME, and feature attribution were prominent, while symbolic reasoning was commonly absent or applied after detection rather than deeply integrated into learning or inference. Edge and edge-cloud deployment received substantial attention, although validation under realistic resource, latency, and communication constraints was more limited. User-centered explainability also remained comparatively underdeveloped, with many explanations oriented toward technical interpretation rather than adaptive support for domain users.

These patterns support the central argument of this paper: trustworthy IoT anomaly detection requires more than accurate detection and post-hoc explanation. The observed limitations in current approaches, including insufficient reasoning integration, limited deployment validation, and inadequate support for human decision making, demonstrate the need for future IoT systems to jointly consider the five core dimensions proposed by our framework.

\subsection{Demonstration of Framework's Design Utility}
\begin{table}[!t]
	\centering
	\caption{Illustrative Capability Profiles for Representative Studies}
	\label{tab:illustrative_trustworthiness_profile}
	\renewcommand{\arraystretch}{1.2}
	\begin{tabular}{
			|>{\fontsize{7}{8}\selectfont\RaggedRight\arraybackslash}p{1.6cm}|
			>{\fontsize{7}{8}\selectfont\RaggedRight\arraybackslash}p{0.36cm}|
			>{\fontsize{7}{8}\selectfont\RaggedRight\arraybackslash}p{0.36cm}|
			>{\fontsize{7}{8}\selectfont\RaggedRight\arraybackslash}p{0.36cm}|
			>{\fontsize{7}{8}\selectfont\RaggedRight\arraybackslash}p{0.36cm}|
			>{\fontsize{7}{8}\selectfont\RaggedRight\arraybackslash}p{0.38cm}|
			>{\fontsize{7}{8}\selectfont\RaggedRight\arraybackslash}p{3.9cm}|
		}

		\hline
		\rowcolor{gray!25}
		\textbf{Study} & \textbf{D1} & \textbf{D2} & \textbf{D3} & \textbf{D4} & \textbf{D5} & \textbf{Framework Interpretation}\\
		\hline
		Rathi and Shandilya \cite{rathiTrustworthyIoTModular2025}
		& 2 & 0 & 2 & 3 & 2 &deployment-aware XAI pipeline, but no symbolic reasoning\\
		\hline
		Kalutharage et al. \cite{kalutharageNeurosymbolicLearningDomain2025}
		& 3 & 2 & 2 & 3 & 2 &  Explanation through cybersecurity knowledge/domain, but symbolic reasoning is largely inference-level\\
		\hline
		Bizzarri et al. \cite{bizzarriNeuroSymbolicArtificialIntelligence2024}
		& 1 & 3 & 3 & 1 & 1 & deeper neural-symbolic integration, but no strong analyst facing explanations or deployment validation\\
		
		\hline
	\end{tabular}
	\vspace{-5mm}	
\end{table}
\label{sec:illustrative_framework_guided_design}
To illustrate the design utility of the proposed framework, we consider its application to an ongoing effort to develop a neuro-symbolic framework for anomaly and medical event detection in multivariate healthcare time-series. This application is not presented as an empirical validation; rather, it demonstrates how the five dimensions can guide the translation of explainability, symbolic reasoning, architectural, deployment, and human-centered requirements into concrete design choices.

\begin{figure*}[!ht]
	\centering
	\centerline{\includegraphics[width=\linewidth]{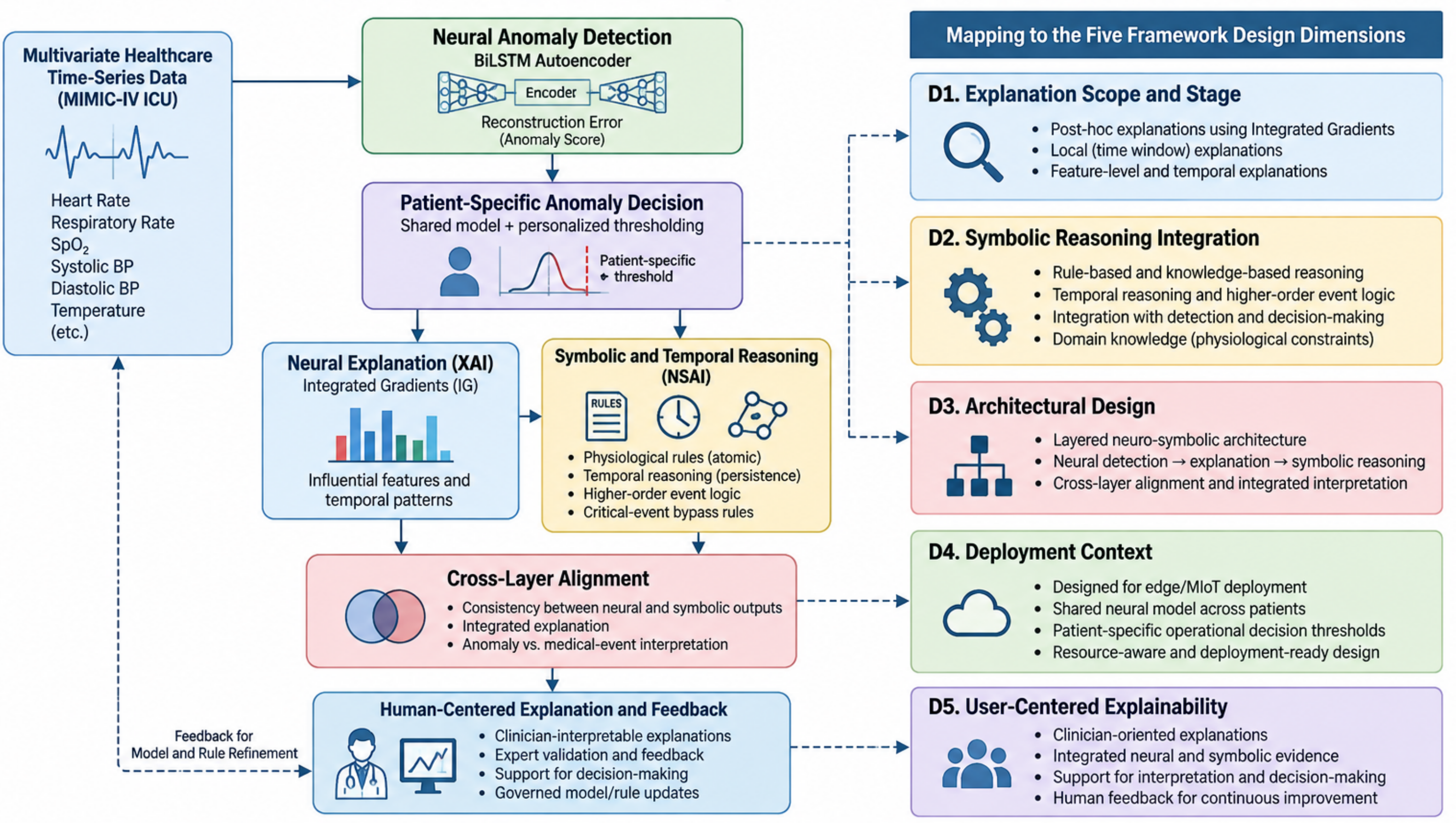}}
	\caption{Illustrative framework guided design for multivariate healthcare time-series anomaly and medical event detection. The five dimensions of the proposed unified XAI--NSAI framework inform design choices. This is merely a design illustration rather than an empirical evaluation.}
	\label{fig:healthcare_framework_guided_design} 
	\vspace{-3.4mm}	
\end{figure*}

As shown in \Cref{fig:healthcare_framework_guided_design}, the design combines neural anomaly detection and patient specific operational thresholds with post-hoc neural explanation, physiological and temporal symbolic reasoning, cross-layer alignment, and human-centered explanation and feedback. These components provide application specific realizations of the five framework dimensions.

The example demonstrates how the proposed dimensions can serve not only as an analytical lens for comparing existing approaches, but also as structured design guidelines for integrating application specific XAI and NSAI mechanisms. The proposed analytical framework does not prescribe the specific mechanisms illustrated in \Cref{fig:healthcare_framework_guided_design}, nor does it imply that these mechanisms constitute the only valid realizations of the five dimensions.

\section{Discussion}
\label{sec:discussion}

Our proposed framework emphasizes that detection performance alone is insufficient to characterize trustworthy IoT anomaly detection systems. By extending evaluation beyond detection performance to complementary design dimensions, the framework enables comparison of heterogeneous approaches and identification of capability gaps and design trade-offs that may otherwise be overlooked. A key strength of the framework is that it provides a multidimensional capability profile rather than reducing system assessment to a single performance measure. This representation makes design trade-offs explicit while avoiding the implication that these heterogeneous characteristics can be reduced to a single measure of trustworthiness. 

The proposed framework also provides a design guidance for future IoT anomaly detection systems. For example, systems intended for edge deployment may require lightweight explanations and efficient reasoning, while systems used in healthcare, industrial monitoring, or cybersecurity may require stronger domain-specific explanations, domain grounded reasoning, and human-in-the-loop feedback.

The framework also highlights several research needs. Future systems should better integrate symbolic reasoning into learning and inference, rather than using rules or knowledge graphs only as post-processing tools. They should also address edge-cloud collaboration, reasoning overhead, latency, and synchronization of symbolic knowledge across distributed IoT environments. In addition, future systems should move beyond static technical explanations toward adaptive explanations that reflect user role, expertise, task context, and operational needs. Standardized evaluation methods are also needed to assess not only detection accuracy, but also explanation faithfulness, reasoning consistency, deployment cost, and usefulness to human decision makers.

\section{Threats to Validity}
\label{sec:validity}

The proposed framework has several limitations. First, the literature analysis used to help in developing the framework was structured but selective; the selected studies are intended to represent important design patterns rather than all XAI-, NSAI-, and hybrid XAI--NSAI-based anomaly detection models. Second, assigning readiness scores across framework dimensions involves interpretive judgment, especially when distinguishing rule-based interpretability from deeper symbolic reasoning integration. Third, the proposed scoring option is qualitative and illustrative, not a validated quantitative trustworthiness metric. Fourth, the framework is evaluated through representative literature mapping rather than through large-scale empirical deployments or studies. Finally, reliable framework based assessment depends on the quality of the underlying evidence and evaluation measures; flawed labels, biased benchmarks, or inappropriate metrics may distort the resulting capability profile.

\section{Conclusion And Future Work}
\label{sec:conclusion}

This paper proposed a unified XAI--NSAI analytical framework for trustworthy IoT anomaly detection. To the best of our knowledge, this is the first framework that provides a structured basis for examining how model explanation and symbolic reasoning interact with architectural, deployment, and user-centered considerations that are often treated independently. The framework's five design dimensions enable a multidimensional characterization of system capabilities, limitations, and design trade-offs rather than reducing assessment to detection performance or explanation availability alone.

Applying the framework to three representative XAI, NSAI, and hybrid approaches illustrates its analytical utility for comparing systems and identifying capability gaps. The analysis highlights complementary strengths and limitations: post-hoc XAI can improve model transparency without providing explicit domain-grounded reasoning, while symbolic and neuro-symbolic approaches can strengthen domain consistency yet still leave gaps in human facing explanation or deployment validation. In addition, a medical event detection illustration demonstrated how the proposed framework can guide the design of an application-specific system by structuring considerations of explainability, symbolic reasoning, deployment, and human-centered requirements.

Overall, the framework provides a common analytical structure for examining capabilities relevant to trustworthy IoT anomaly detection without treating any individual design choice or qualitative score as a measure of trustworthiness itself. Future work will focus on validating the framework with domain experts and through real-world IoT deployments, refining the optional qualitative readiness scoring approach, and investigating standardized evaluation criteria for explanation usefulness and faithfulness, reasoning quality and consistency, deployment cost and feasibility, and utility to human decision makers.


\bibliographystyle{IEEEtran}
\bibliography{bibfile}




\end{document}